\RequirePackage{fix-cm}

\documentclass[twocolumn]{svjour3}          

\smartqed  
\usepackage[misc]{ifsym}
\usepackage{ragged2e}
\usepackage{graphicx}
\usepackage{amsmath}
\usepackage{amssymb}
\usepackage{algorithmic}
\usepackage{array}
\usepackage{dblfloatfix}
\usepackage{url}
\usepackage[caption=false,font=footnotesize,labelfont=sf,textfont=sf]{subfig}
\usepackage{multirow}
\usepackage[sort&compress,numbers]{natbib}
\usepackage[pagebackref=true,breaklinks=true,letterpaper=true,colorlinks,bookmarks=false]{hyperref}
\usepackage{booktabs}
\usepackage{mathptmx} 
\usepackage{color}
\renewcommand{\raggedright}{\leftskip=0pt \rightskip=0pt plus 0cm}

\begin{document}

\title{2D Human Pose Estimation: A Survey}

\author{Haoming Chen \textsuperscript{1\ddag}\thanks{\ddag \space The first two authors have equal contribution.}  \and
  Runyang Feng  \textsuperscript{1\ddag}\thanks{* Corresponding Author.}  \and
  Sifan Wu \textsuperscript{1}  \and
  Hao Xu \textsuperscript{2}  \and
  Fengcheng Zhou \textsuperscript{1*}  \and
        Zhenguang Liu \textsuperscript{1}
}


\institute{\Letter\quad 
            Fengcheng Zhou\\  \vspace{0.4em}
            \hspace*{0.48cm} zfc@zjsu.edu.cn
 \at
 {\textsuperscript{1}} \hspace*{0.25cm}Zhejiang Gongshang University, Hangzhou, China. \\
 {\textsuperscript{2}} \hspace*{0.25cm}Zhejiang Lab, Hangzhou, China.
}

\date{Received: date / Accepted: date}
\maketitle
\begin{abstract}
Human pose estimation aims at localizing human anatomical keypoints or body parts in the input data (\emph{e.g.}, images, videos, or signals). 
It forms a crucial component in enabling machines to have an insightful understanding of the behaviors of humans, and has become a salient problem in computer vision and related fields.
Deep learning techniques allow learning feature representations directly from the data, significantly pushing the performance boundary of human pose estimation.
In this paper, we reap the recent achievements of 2D human pose estimation methods and present a comprehensive survey.
Briefly, existing approaches put their efforts in three directions, namely \emph{network architecture design}, \emph{network training refinement}, and \emph{post processing}.
Network architecture design looks at the architecture of human pose estimation models, extracting more robust features for keypoint recognition and localization.
Network training refinement tap into the training of neural networks and aims to improve the representational ability of models.
Post processing further incorporates model-agnostic polishing strategies to improve the performance of keypoint detection.
More than $200$ research contributions are involved in this survey, covering methodological frameworks, common benchmark datasets, evaluation metrics, and performance comparisons.
We seek to provide researchers with a more comprehensive and systematic review on human pose estimation, allowing them to acquire a grand panorama and better identify future directions.
\end{abstract}

\section{Introduction}\label{intro}
\begin{figure*}[t]
\begin{center}
\includegraphics[width=1\linewidth]{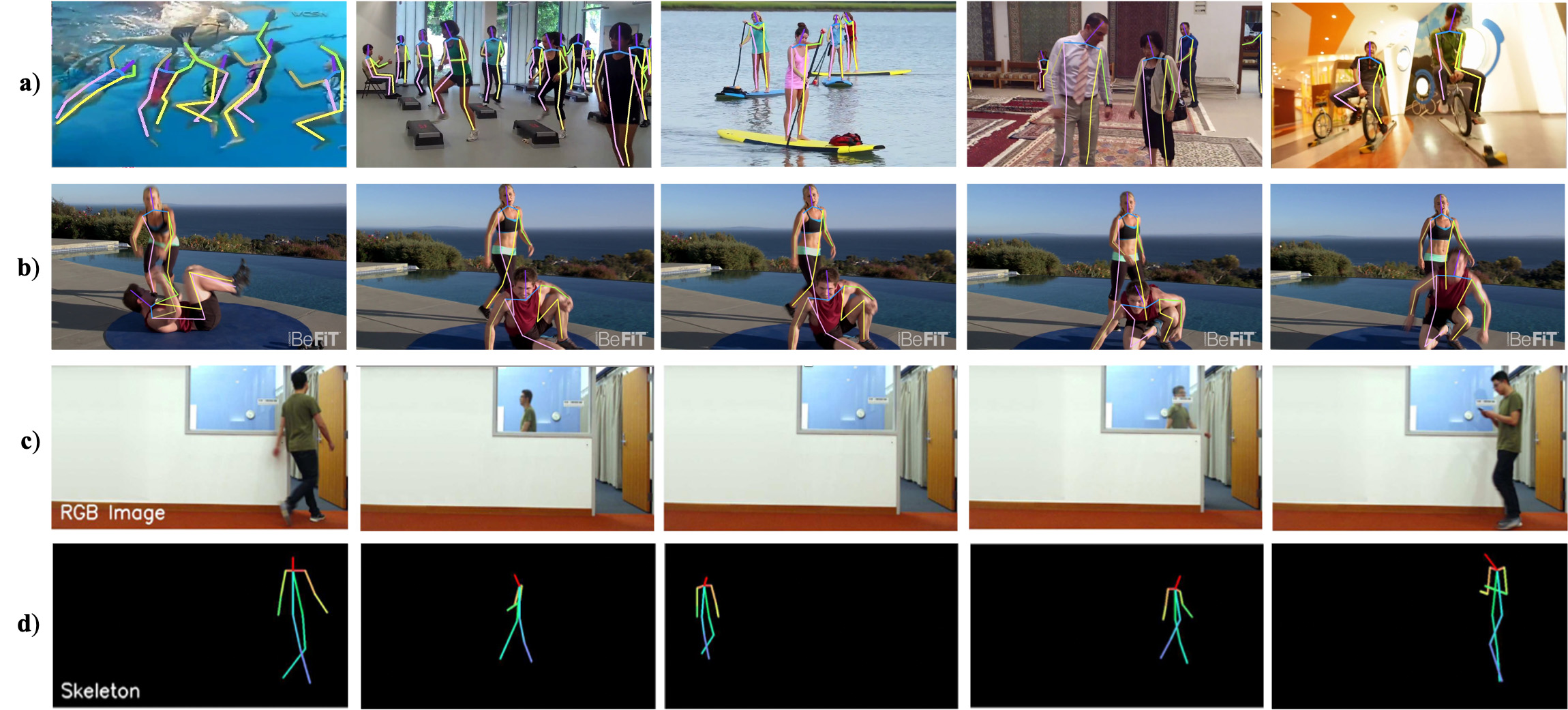}
\end{center}
\caption{An illustration of 2D human pose estimation on multimedia data, including images a), videos b), and RF signals c), d). Note that RGB images c) are presented for visual reference of RF signals-based HPE, and d) shows the skeleton extracted from the RF signals \emph{alone}. The pictures in c) and d) are cited from \cite{zhao2018through}.}
\label{fig:examples}
\end{figure*}

As a compelling and fundamental problem in computer vision, human pose estimation (HPE) has attracted intense attention in recent years.
As shown in Fig. \ref{fig:examples}, the goal of 2D HPE is to: 1) recognize different person instances within the multimedia data (RGB images, videos, RF signals, or radar) recorded by sensors, and 2) to localize a set of pre-defined human anatomical keypoints for each person.
\textcolor{black}{As the cornerstone of human-centric visual understanding, 2D HPE provides the groundwork for tackling multitudinous higher-order computer vision tasks such as 3D human pose estimation \cite{martinez2017simple, chen20173d, mehta2017vnect, zeng2021learning, zou2021eventhpe, garau2021deca, wehrbein2021probabilistic, gao2016iterative, gao2015user}, human action recognition \cite{wang2013action, baccouche2011sequential, ji20123d}, human parsing \cite{ruan2019devil, gong2018instance, gong2017look}, pose tracking \cite{wang2020combining, girdhar2018detect, xiao2018simple}, motion prediction \cite{liu2019towards, Our, liu2021aggregated}, human motion retargeting \cite{chan2019everybody, kappel2021high, naksuk2005whole}, and vision-and-language conversion  \cite{guo2021adavqa, Datta_2019_ICCV, mogadala2021trends, vqa2, vqa3}.}
HPE supports a wide spectrum of applications including human behaviors understanding, motion capture, violence detection, crowd riot scene identification, human-computer interaction, and autonomous driving.
\par
\raggedright
Earlier methods \cite{wang2008multiple, wang2013beyond, zhang2009efficient, sapp2010cascaded} adopt the probabilistic graphical model to represent relations between joints.
Unfortunately, these methods rely heavily on hand-crafted features which limit their generalization and performance.
More recently, the deep learning techniques \cite{lecun1998gradient, schmidtke2021unsupervised, liu2022convnet, shang2019annotating, li2021interventional} enable learning feature representations automatically from data, which has significantly contributed to the advancement of human pose estimation.
These deep learning-based approaches \cite{Toshev_2014_CVPR, liu2021deep, xiao2018simple, sun2019deep, Cao_2017_CVPR, liu2021combining, li2021human, li2021online}, commonly building upon the success of convolutional neural networks, have achieved outstanding performance on this task.
Given the rapid development, this paper seeks to track recent progress and summarize their accomplishments to deliver a clearer panorama for 2D human pose estimation.
\par
Several excellent surveys related to human pose estimation have been published, as presented in the Table \ref{surveys}, involving studies in areas of human motion capture and analysis \cite{moeslund2001survey, moeslund2006survey, poppe2007vision, ji2009advances}, activity recognition and 2D/3D HPE \cite{zheng2020deep, chen2020monocular, liu2015survey}, \emph{etc}. 
However, few surveys are dedicated to  2D human pose estimation.
On the other hand, most of existing surveys cast existing approaches into \emph{single-person}  and \emph{multi-person} pose estimation methods. 
The \emph{single-person pose estimators} typically focus on the model architectures for keypoint detection, and can perform well in the multi-person pose estimation scenarios by predicting pose for each individual person within his/her bounding box.
Therefore, a pose estimation model can accommodate both \emph{single-person} and \emph{multi-person} scenes, and the division as above might be unnecessary.
Moreover, while human pose estimation on images or videos has been widely concerned,  to the best of our knowledge there is still no work that summarizes signal-based human pose estimation, \emph{e.g.}, RF signals and radar signals.
\par 
In this paper, we roughly cast human pose estimation methods into three categories, each containing several subcategories on a finer level.
(1) \emph{Network architecture design} approaches attempt to devise vigorous models that capture robust representations across different scenes to effectively detect keypoints.
Methods in this category concentrate on extracting and processing human body features within a person bounding box \cite{xiao2018simple, fang2017rmpe} or over the entire image \cite{Cao_2017_CVPR, cheng2020higherhrnet}.
(2) \emph{Network training refinement} approaches aim at optimizing neural network training, trying to improve the model ability without changing the network structure. Towards this aim, they engage in data augmentation techniques \cite{wang2021human, bin2020adversarial}, model training strategies \cite{xia2017joint, nie2018human}, loss function constraints \cite{zhou2020occlusion, chen2018cascaded}, and domain adaption methods \cite{hidalgo2019single, xu2020alleviating}. 
(3)  \emph{Post processing} methods focus on pose polishment upon the coarse pose estimates to improve the performance.
The methods within this category usually behave as a model-agnostic plugin. Representative techniques for pose polishement include quantization error minimization \cite{zhang2020distribution, huang2020devil} and pose resampling \cite{liu2021deep, wang2020combining}. 
\emph{Furthermore}, we also discuss the rarely involved topic of reconstructing 2D human poses from signals such as RF signals \cite{zhao2018through, wang2019can} and radar signals \cite{li2020capturing}, hoping to fill the knowledge gap.
\renewcommand\arraystretch{1.7}
\begin{table*}
\vspace{0.2em}\caption{Summary of previous surveys and reviews related to the human pose estimation.}\label{surveys}
  \resizebox{1\textwidth}{!}{
  \begin{tabular} {p{8.5cm} p{1.0cm} p{1.0cm} p{10.0cm} p{1.0cm} p{1.0cm}} %
  \hline
   Survey Title 	&Year 	&Venue 	&Content &\textcolor{black}{Single-Person} &\textcolor{black}{Multi-Person}\\ 
	\hline
A Survey of Computer Vision-Based Human Motion Capture \cite{moeslund2001survey} &2001 &CVIU &A survey of different functionalities in motion capture system, including initialization, tracking, pose estimation, and recognition. & \textcolor{black}{\checkmark} &\textcolor{black}{\checkmark} \\
	\hline
A survey of advances in vision-based human motion capture and analysis \cite{moeslund2006survey} &2006 &CVIU & A survey of advances in human motion capture and analysis from 2000 to 2006. & \textcolor{black}{\checkmark} &\textcolor{black}{\checkmark}\\
	\hline
Vision-based human motion analysis: An overview \cite{poppe2007vision} &2007 &CVIU & An overview of markerless vision-based human motion analysis. & \textcolor{black}{\checkmark} &\textcolor{black}{\checkmark}\\
	\hline
Advances in view-invariant human motion analysis: A review \cite{ji2009advances} &2010 &TSMCS & A review of major issues in human motion analysis system, including human detection, view-invariant pose representation and estimation, and human behavior understanding. &\textcolor{black}{\checkmark} &\textcolor{black}{\checkmark} \\
	\hline
Visual analysis of humans \cite{moeslund2011visual} &2011 &Book & A comprehensive overview of human analysis such as pose estimation and applications. &\textcolor{black}{\checkmark} &\textcolor{black}{\checkmark} \\
	\hline
Human pose estimation and activity recognition from multi-view videos: Comparative explorations of recent developments \cite{holte2012human} &2012 &JSTSP & A review of multi-view based 3D human pose estimation and activity recognition. & \textcolor{black}{\checkmark} & \\
    \hline
A survey of human pose estimation: the body parts parsing based methods \cite{liu2015survey} &2015 &JVCIR & A survey of human parsing based 2D/3D human pose estimation. & \textcolor{black}{\checkmark} &\textcolor{black}{\checkmark} \\
	\hline
Human pose estimation from monocular images: A comprehensive survey \cite{gong2016human} &2016 &Sensors &A survey of conventional and  deep learning methods for human pose estimation. & &\textcolor{black}{\checkmark}\\
	\hline
3d human pose estimation: A review of the literature and analysis of covariates \cite{sarafianos20163d} &2016 &CVIU &A review of the advances in 3D human pose estimation from RGB images or image sequences. & \textcolor{black}{\checkmark} &\\
	\hline
Monocular human pose estimation: a survey of deep learning-based methods \cite{chen2020monocular} &2020 &CVIU &A survey of monocular based 2D/3D human pose estimation employing deep learning methods. & \textcolor{black}{\checkmark} &\textcolor{black}{\checkmark}\\
	\hline
The progress of human pose estimation: a survey and taxonomy of models applied in 2D human pose estimation \cite{munea2020progress} &2020 & IEEE Access &A survey of researches on 2D human pose estimation. & \textcolor{black}{\checkmark} &\textcolor{black}{\checkmark}\\
	\hline
Deep learning-based human pose estimation: A survey \cite{zheng2020deep} &2020 &arXiv &A survey of deep learning-based 2D/3D human pose estimation. &\textcolor{black}{ \checkmark} &\textcolor{black}{\checkmark}\\
	 \hline
    \end{tabular}}
\end{table*}

\vspace{-1em}
\subsection{Scope}
Our scope is limited to 2D human pose estimation with deep learning, we do not consider the conventional non-deep-lear-ning methods. 
Topics such as the applications of 2D HPE \cite{zheng2020deep} and the representations of human body models \cite{chen2020monocular} that have been adequately covered by other reviews will not be detailed here either.
Nevertheless, there are still a breathtaking number of papers on 2D HPE, hence it is necessary to establish a selection criterion, in such a way that we restrict our attention to the top journal and conference papers since $2014$.
In light of these constraints, we sincerely apologize to those authors whose works are not incorporated into this paper. 
\subsection{Outline}
The rest of this paper is organized as follows.
In Section \ref{problem} we provide problem formulations for 2D human pose estimation, and briefly discuss  the technological challenges of 2D HPE.
Then, we present works on network architecture design in Section \ref{model}, introduce network training refinement methods in Section \ref{train}, and review post processing approaches in Section \ref{post}.
Subsequently, we summarize the common benchmark datasets, evaluation metrics, and performance comparisons in Section \ref{dataset}.
We further provide discussions in Section \ref{discussion}, including open questions, signal-based 2D HPE, and future research directions.
Finally, we conclude the paper in Section \ref{conclusion}.
\section{Problem Statement}\label{problem}
In this section, we first define the problem of 2D HPE on the image and video data, followed by the discussion of technological challenges in this task.
\subsection{The Problem}
Formally, the human pose estimation problem can be formualted as follows.  
Given an image or a video as input, the goal is to detect the \textbf{poses} of all persons in the input data. 
\textcolor{black}{Technically, presented with an observed image ${I}$, we aim to detect the pose $ $ of each person $i$ in the image $\mathbf{P} = \{\mathbf{P}_{i}\}_{i=1}^n$, where $n$ denotes the number of persons in ${I}$.}

\textcolor{black}{To describe human poses, skeleton-based model \cite{felzenszwalb2005pictorial}, contour-based model \cite{ju1996cardboard}, and volume-based model \cite{sidenbladh2000framework} have been proposed in previous works.}
In particular, the contour-based representation contains rough body contour and limb width information while the volume-based representation describes 3D human shapes.
The skeleton-based model, which characterizes the human body as a set of pre-defined joints, has been widely employed in 2D HPE. 
\subsection{Technical Challenges}
Ideally, an algorithm that is both highly accurate and efficient is desired to solve the problem of 2D HPE.
High accuracy detection ensures a precise human body information to facilitate downstream tasks such as 3D HPE and action recognition, while high efficiency allows real-time computing in different devices such as desktops and mobile phones.

Challenges in \emph{accurate} pose detection come from several aspects.
(1) Nuisance phenomena such as under/over-exposure and human-objects entanglement frequently occur in real-world scenes, which may easily lead to detection failure.
(2) Due to the highly flexible human kinematic chains, pose occlusions even self-occlusions in many scenarios are inevitable, which will further confuse keypoint detectors using visual features. 
(3) Motion blur and video defocus do frequently happen in videos, which deteriorates the accuracy of pose detection.

When the pose estimation algorithms are applied to practical applications, besides accurate estimation, the running speed (\emph{efficiency}) is also important.
However, high accuracy and high efficiency are often in conflict to each other since the high accuracy models tend to be deeper, requiring increased resources for computation and storage. 
For example, HRNet-W48 \cite{sun2019deep} has achieved state-of-the-art results on multiple benchmarks, which however has difficulties in achieving real-time pose estimation even with the help of powerful NVIDIA GTX-1080TI GPUs. Consequently, light-weight models with comparable precision are much coveted for mobile or wearable devices.
\section{Network Architecture Design Methods}\label{model}
A key advantage of modern deep learning methods is the ability to learn feature representations automatically from data.
However, feature quality is closely related to the network architecture, therefore the topic of network design deserves to be investigated deeply.
Correspondingly,  \emph{network architecture design} methods aim at extracting powerful features by investigating various network designs to address human pose estimation. 
In this section, we set out to introduce these approaches in detail with a focus on their network architectures.
\begin{figure*}[t]
\begin{center}
\includegraphics[width=1\linewidth]{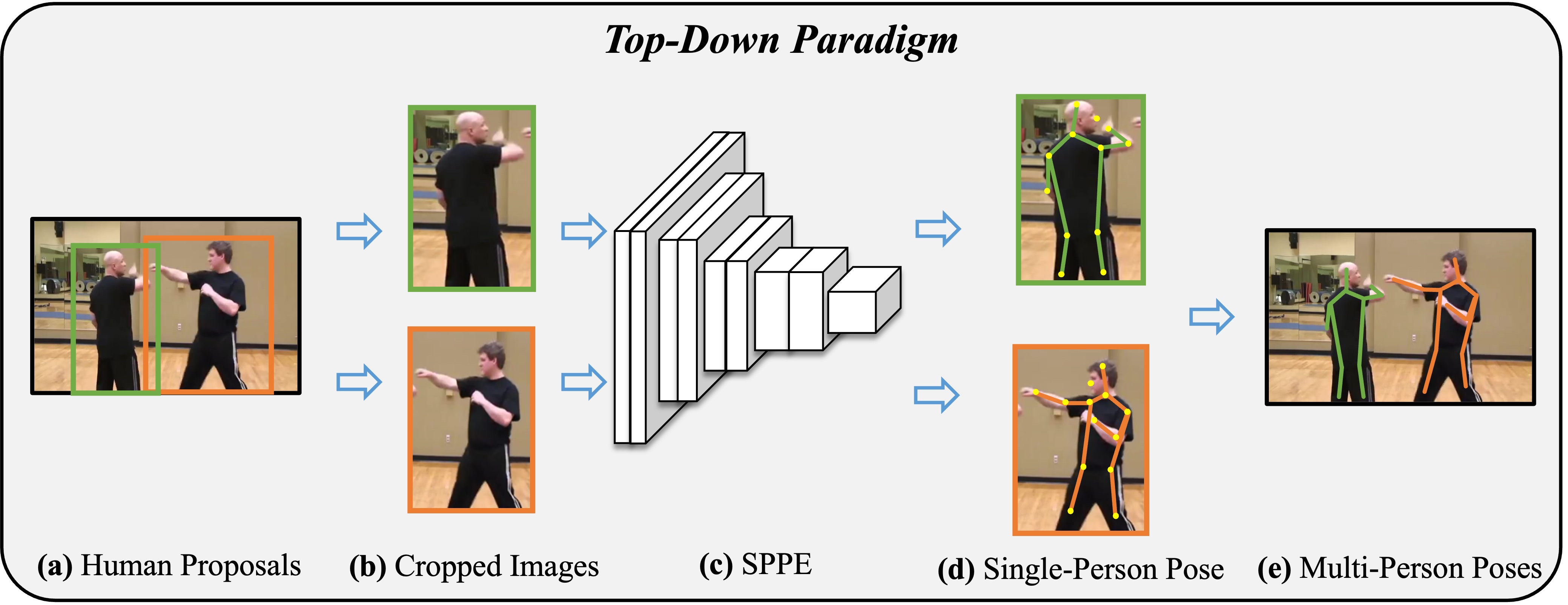}
\end{center}
\caption{A classical pipeline of top-down framework for human pose estimation. (a): Original image in the dataset. The goal is to detect the poses of all persons in the input image. An off-the-shelf object detector is  employed to perform person detection and gives the human proposals. (b): The regions of human proposals are cropped from the original image to form the single person images. 
    (c) Each cropped image is subjected to single person pose estimation (SPPE) to obtain the estimated pose, which is illustrated in (d). (e): All estimated poses are projected to the original image and yield the final results.}
\label{fig:top-down}
\end{figure*}
\par
On a high level, these approaches typically fall into two general frameworks, namely \textbf{top-down} framework \cite{liu2021deep, Wei_2016_CVPR, fang2017rmpe, newell2016stacked, bertasius2019learning, sun2019deep} and \textbf{bottom-up} framework \cite{Cao_2017_CVPR, kreiss2019pifpaf, geng2021bottom, luo2021rethinking, wei2020point, jin2020differentiable}.
The top-down paradigm employs a two-step procedure that first detects human bounding boxes and then performs single person pose estimation for each bounding box, which is exemplified in Fig. \ref{fig:top-down}. 
The bottom-up paradigm adopts the part-based procedure that first locates identity-free keypoints and then groups them into different person instances.
We may further divide different methods in these two paradigms into fine-grained sub-categories, where the top-down approaches are categorized into \emph{regression-based} \cite{Toshev_2014_CVPR, carreira2016human}, \emph{heatmap-based} \cite{sun2019deep, xiao2018simple}, \emph{video-based} \cite{liu2021deep, luo2018lstm}, and \emph{model compressing-based} \cite{zhang2019fastpose, yu2021lite} methods, and the bottom-up approaches are classified into \emph{one stage} \cite{2020Single, geng2021bottom} and \emph{two-stage} methods \cite{Cao_2017_CVPR, kreiss2019pifpaf}. 
In what follows, we introduce these categories in detail.
\subsection{Top-Down Framework}
\subsubsection{Regression-Based Methods}\label{regression}
Earlier works \cite{zhang2020key, wang2020graph, qiu2020peeking, zhang2018poseflow, sun2018integral, fieraru2018learning, sun2017compositional, carreira2016human, fan2015combining, Toshev_2014_CVPR, li2014heterogeneous} attempt to learn a mapping from input image to the pre-defined kinematic joints via an end-to-end network, and directly regress the keypoint coordinates, which we refer to as the \emph{regression-based} approaches.
\par
For instance, \textcolor{black}{DeepPose \cite{Toshev_2014_CVPR} sets the precedent of human pose estimation with deep learning technique. It \cite{Toshev_2014_CVPR} first employs an iterative architecture to extract image features with the cascaded convolutional neural networks (AlexNet \cite{krizhevsky2012imagenet}), and subsequently regresses the joint coordinates with fully connected layers.
Inspired by the remarkable performance of deep learning works such as DeepPose, researchers gradually turned from conventional methods to the deep learning ones.}
Building upon the GoogleNet \cite{szegedy2015going}, \cite{carreira2016human} proposes a self-correcting model, which progressively changes the initial joint coordinates estimations instead of {directly predicting joint positions}.
\cite{sun2017compositional} presents a structure-aware regression approach that utilizes a novel re-parameterized pose representation of bones. This method is constructed on the ResNet50 \cite{he2016deep}, and is able to capture more structural human body information such as joint connections, which enriches the pure joint-based pose descriptions.
\par
\textcolor{black}{Graph convolutional network (GCN) \cite{kipf2016semi} has recently been widely explored, which employs nodes and edges to represent entities and their correlations. Upon convolutions on the graph,  the feature of a node is enhanced by incorporating features from the neighboring nodes.  Compared to traditional methods, GCN provides another competitive and novel model to characterize the human body.}
\cite{qiu2020peeking} casts the human body as a graph structure where the nodes represent joints and the edges represent bones, and proposes to estimate invisible joints using an Image-Guided Progressive GCN module. 
\par
\textcolor{black}{Attention mechanism has greatly advanced the representation learning, and the Transformer \cite{carion2020end, jaderberg2015spatial, vaswani2017attention, zhu2020deformable} built upon self-attention has established  new state-of-the-arts on multiple visual understanding tasks such as object detection, image classification, and semantic segmentation.}
\cite{li2021pose} presents a cascaded Transformers performing end-to-end regression of human and keypoint detection, which first detects the bounding boxes for all persons and then separately regresses all joint coordinates for each person.

\par
\textcolor{black}{The regression-based methods are highly efficient and show promising potential in real-time applications. Unfortunately, such approaches directly output a single 2D coordinates for each joint, failing to consider the area of the body part.}
To tackle this issue, heatmap-based approaches are introduced, which localize the keypoints by probabilistic heatmaps instead of determined coordinates.
\subsubsection{Heatmap-Based Methods}\label{heatmap}
In order to overcome the shortcomings of direct coordinate regression, heatmap-based joint representations have been widely adopted \cite{pfister2015flowing}, which leads to an easier optimization and a more robust generalization.
Specifically, the heatmap $H_i$ is generated via a 2D Gaussian centered at each joint location $( x_i, y_i)$, encoding the probability of the location being the $i^{th}$ joint.
During training, the goal is to predict $N$ heatmaps $\{H_1, H_2, .., H_N\}$ for a total of $N$ joints.
Representative \emph{heatmap-based} approaches include:
\par
\begin{figure*}[t]
\begin{center}
\includegraphics[width=1\linewidth]{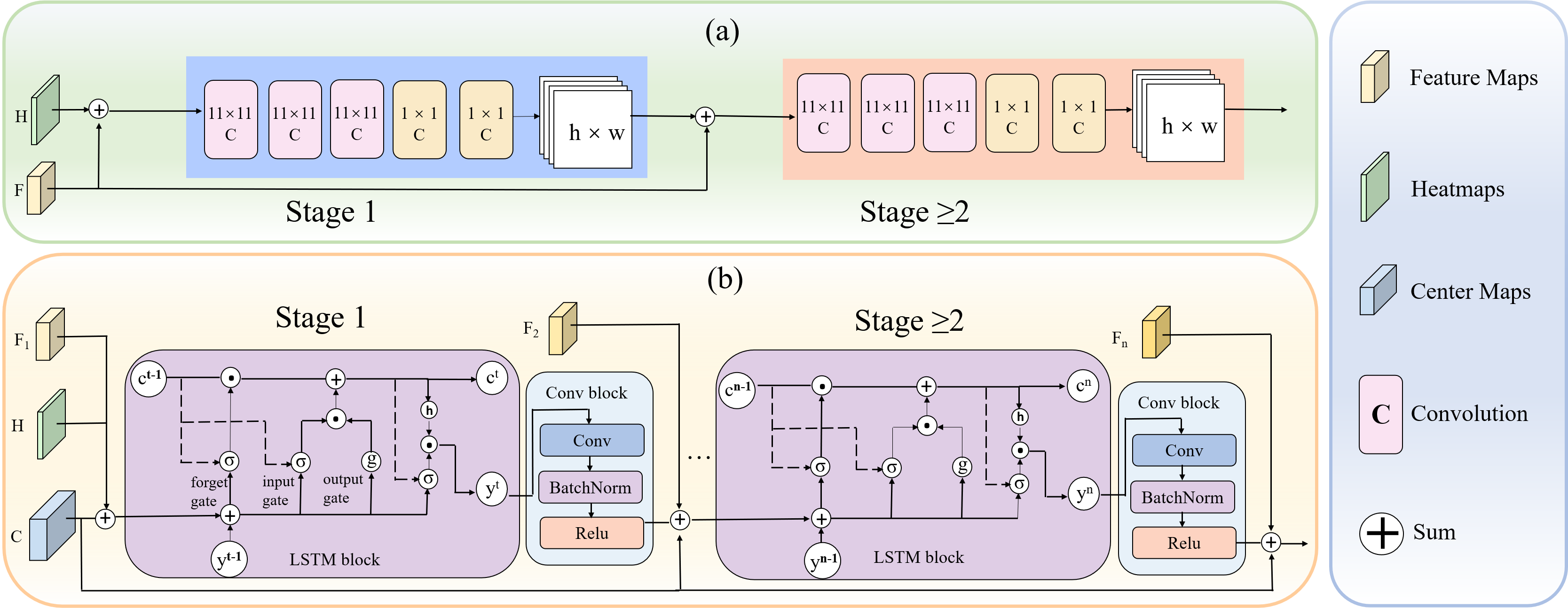}
\end{center}
\caption{Illustration of the networks based on iterative architecture. The top portion (a) of the figure depicts the structure of Convolutional Pose Machine  \cite{Wei_2016_CVPR} while the bottom part (b) shows the network of LSTM Pose Machines \cite{luo2018lstm}. In \cite{Wei_2016_CVPR}, the prediction of each stage and image features are concatenated for the subsequent stage. \cite{luo2018lstm} extends \cite{Wei_2016_CVPR} with LSTM. The heatmaps predicted at the previous stage, frame features, and a center map are concatenated to fed into the subsequent stage. Note that different stages in \cite{Wei_2016_CVPR} aim at optimizing pose estimation of the same image, while different stages in \cite{luo2018lstm} process various video frames. }
\label{fig:iter}
\end{figure*}

 
\textcolor{black}{\textbf{Iterative Architecture}\quad}
Conventionally, the iterative architecture \cite{Toshev_2014_CVPR, ramakrishna2014pose, Wei_2016_CVPR, luo2018lstm, carreira2016human} is designed to produce and refine the keypoint heatmaps.
\cite{ramakrishna2014pose} presents an inference machine model which gradually infers the locations of joints in multiple stages.
\cite{Wei_2016_CVPR} further extends the architecture of \cite{ramakrishna2014pose} and builds a sequential prediction framework, which employs sequential convolutions to implicitly model long-range spatial dependencies between human body parts.
This approach harvests increasingly refined estimates for joint locations by operating on the results of previous stage, as shown in Fig. \ref{fig:iter}. 
\cite{Wei_2016_CVPR} additionally proposes intermediate supervision to alleviate the inherent problem of \emph{vanishing gradients} in the iterative architectures.
\par
Although the intermediate supervision strategy relieves the \emph{vanishing gradients} of multi-stage models, each stage still fails to build a deep sub-network to extract effective semantic features, which greatly limits their fitting capabilities.
This issue has been tackled with the emergence of residual network (ResNet) \cite{he2016deep}, which introduces a shortcut and allows the errors at deeper layers to be back-propagated.
Benefiting from such a way, numerous large models \cite{newell2016stacked, chu2017multi, yang2017learning, liu2018cascaded, tang2018deeply, ke2018multi, chen2018cascaded, xiao2018simple, sun2019deep, su2019multi, cai2020learning, jiang2020pay} have been devised, which greatly boost the process of 2D HPE.
\begin{figure*}[t]
\begin{center}
\includegraphics[width=1\linewidth]{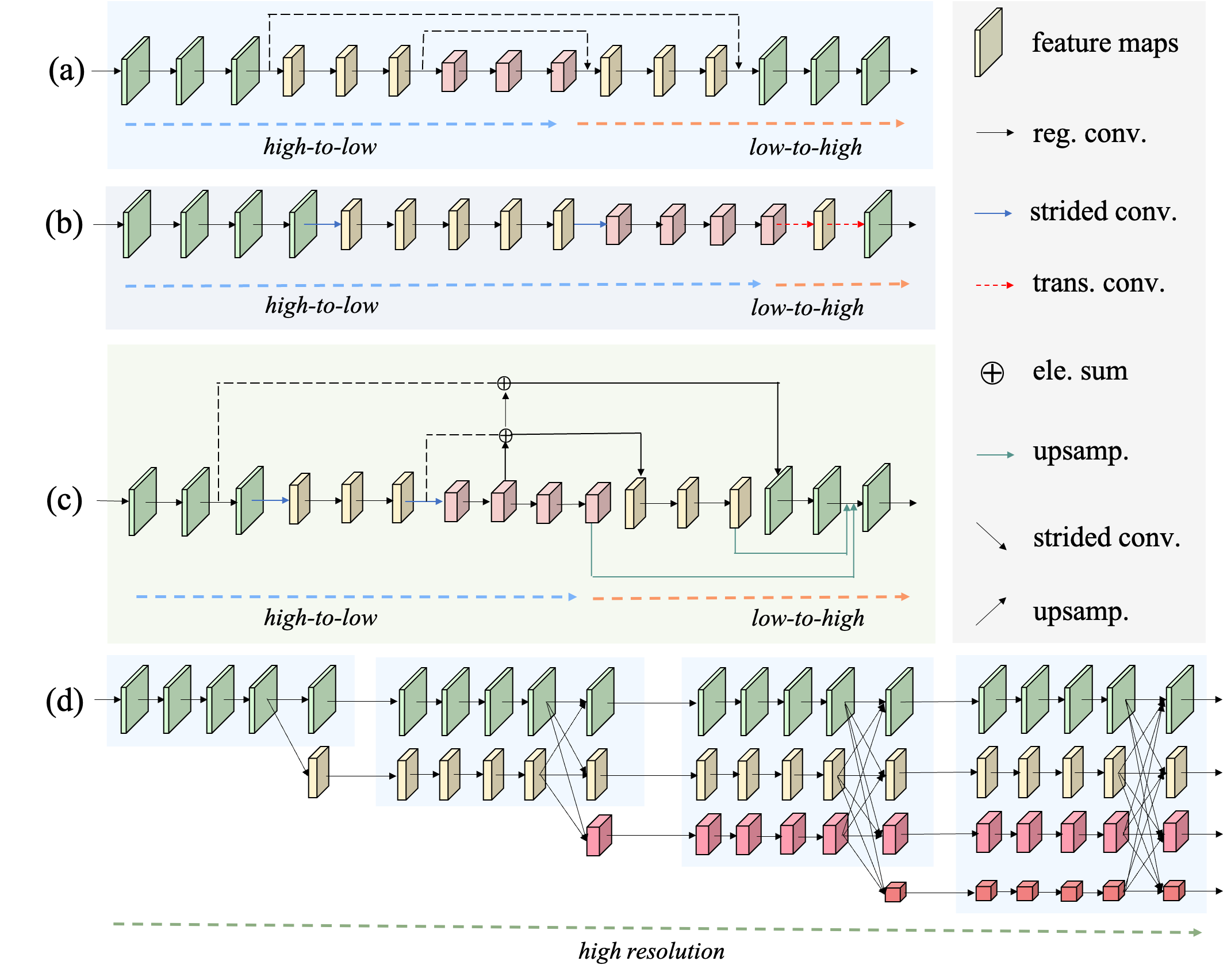}
\end{center}
\caption{Illustration of the classical human pose detector that relys on the \emph{high-to-low} and \emph{low-to-high} framework. (a) Stacked hourglass network \cite{newell2016stacked}. (b) Cascaded pyramid networks \cite{chen2018cascaded}. (c) SimpleBaseline \cite{xiao2018simple}. (d) HRNet \cite{sun2019deep}. Legend: \emph{reg. conv.} = regular convolution layer, \emph{strided conv.} = strided convolution layer for learnable downsampling, \emph{trans. conv.} = transposed convolution layer for learnable upsampling, \emph{ele. sum} = element-wise summation. For the architecture of (a) stacked hourglass, the high-to-low and low-to-high network architectures are symmetric. In (b) and (c), the high-to-low process is performed by a large visual backbone network (ResNet) which is \textbf{heavy}, while the low-to-high process is implemented by some transposed convolutions or directly upsampling, which is \textbf{light}. In (c), the skip-connection (dashed lines) aims to fuse the features with same spatial size in the high-to-low and low-to-high process. In HRNet (d), the high resolution representation is maintained in the entire propagation, and repeated multi-scale fusions are performed, each resolution features receive rich information from all resolutions.
}
\label{fig:net}
\end{figure*}
\par
\textcolor{black}{\textbf{Symmetric Architecture}\quad}
The deep models generally employ a \emph{high-to-low} (downsampling) and \emph{low-to-high} (upsampling) framework, where \emph{high} and \emph{low} denote the resolution of feature representations.
\cite{newell2016stacked} proposes a novel stacked hourglass architecture based on the successive steps of pooling and upsampling, which incorporates features across all scales to capture the various spatial relationships between joints. The stacked hourglass architecture is depicted in Fig. \ref{fig:net} a). Several variations \cite{chu2017multi, yang2017learning, ke2018multi, cai2020learning} that built upon the success of this stacked hourglass architecture are subsequently developed. Specifically, \cite{chu2017multi} extends \cite{newell2016stacked} to Hourglass Residual Units with a side branch including filters with larger receptive field, which greatly increases the receptive fields of the network and automatically learns features across different scales. 
\cite{yang2017learning} further replaces the residual blocks in the stacked hourglass \cite{newell2016stacked} with the Pyramid Residual Modules which enhances the scale invariance of networks.
\cite{ke2018multi} proposes a multi-scale supervision that combines the keypoint heatmaps across all scales, which leads to acquiring abundant contextual features and improves the performance of stacked hourglass network.
\cite{cai2020learning} designs a stacked hourglass-like network, \emph{i.e.}, Residual Steps Network which aggregates features with the same spatial size to produce the delicate localized descriptions.
\cite{tang2019does} employs the hourglass network \cite{newell2016stacked} as backbone, and proposes a part-based branching network to learn the representations specific to different part groups. 
These hourglass-based models retain \emph{symmetric} architecture between high-to-low and low-to-high convolutions. 
\par
\textcolor{black}{\textbf{Asymmetric Architecture}\quad}
Another line of work exploits an \emph{asymmetric} architecture \cite{chen2018cascaded, xiao2018simple, insafutdinov2016deepercut}, where the high-to-low process is \emph{heavy} and the low-to-high process is \emph{light}.
\cite{chen2018cascaded} proposes a Cascaded Pyramid Network (Fig. \ref{fig:net} c) that detects the simple keypoints with a GlobalNet, and handles the difficult keypoints with a RefineNet.
Specifically, the RefineNet consists of several regular convolutions, integrating all levels of feature representations from the GlobalNet.
\cite{xiao2018simple} extends the ResNet \cite{he2016deep} by adding a few deconvolutional layers instead of feature map interpolation, which is depicted in Fig. \ref{fig:net} b).
These methods employ a sub-network of classical classification networks (VGGNet \cite{simonyan2014very} and ResNet \cite{he2016deep}) for \emph{high-to-low} convolution and adopt simple networks for \emph{low-to-high} convolution.
Undoubtedly, such asymmetric network architectures suffer from imbalances in feature encoding and decoding, which potentially affects model performance.
\par
\textcolor{black}{\textbf{High Resolution Architecture}\quad}
\textcolor{black}{Unlike previous models, \cite{sun2019deep} proposes a representative network,   HRNet\footnote{Link of HRNet Project: \url{https://github.com/leoxiaobin/deep-high-resolution-net.pytorch}} (Fig. \ref{fig:net} d), which is able to maintain high resolution representations through the whole process, achieving state-of-the-art results on multiple vision tasks.
This work demonstrates the superiority of high-resolution representations for human pose estimation and inspires a wide spectrum of later researches \cite{jiang2020pay, wang2020combining, liu2021deep}.}
\cite{jiang2020pay} takes  HRNet as the backbone network, and further incorporates the gating mechanism as well as feature attention module to select and fuse discriminative and attention-aware features.

\textcolor{black}{\textbf{Composed Human Proposal Detection}\quad}
\textcolor{black}{The above models concentrate on pose estimation on a given human proposal which is cropped from the entire image, and simply employ off-the-shelf human proposal detectors for proposal identification.  Existing work \cite{fang2017rmpe, li2019crowdpose} has demonstrated that the quality of human proposals (\emph{e.g.}, human position and redundant detection) significantly affects the results of pose estimators.} Therefore, a group of researches direct their efforts in refining human proposals.  For instances,
\cite{papandreou2017towards} presents a multi-person pose estimation method, which employs the Faster-RCNN \cite{ren2015faster} as person detector and the ResNet-101 \cite{he2016deep} as pose detector, and additionally proposes a novel keypoint NonMaximum-Suppression (NMS) strategy to address the problem of pose redundancy.
\cite{fang2017rmpe} utilizes the SSD-512 \cite{liu2016ssd} as human detector and the stacked hourglass \cite{newell2016stacked} as single person pose detector, and further proposes a symmetric spatial transformer network to extract a high-quality single person region from an inaccurate bounding box to facilitate human pose estimation.
\cite{li2019crowdpose} notices that single person bounding boxes in crowded scenes tend to contain multiple people, which deteriorates the performance of the pose detector. 
To tackle this problem, \cite{li2019crowdpose} leverages a joint-candidate pose detector to predict the  heatmaps with multiple peaks, and uses a graph network to perform global joints association.
\par
In contrast, another group of researches propose to perform proposal detection and pose detection jointly. 
\cite{varamesh2020mixture} develops a mixture model which simultaneously infers the human bounding boxes and keypoint locations in a dense regression fashion.
\cite{wei2020point} introduces a template offset model which first gives a good initialization for the human bounding boxes and poses, and then regresses the offsets between initialization and corresponding labels.
\cite{kocabas2018multiposenet} presents a MultiPoseNet which first detects the keypoints and human proposals separately, and then employs a Pose Residual Network to assign the detected keypoints to different bounding boxes. Specifically, the Pose Residual Network is implemented by a residual multilayer perceptron. 
\cite{mao2021fcpose} designs a pose estimation framework, which incorporates dynamic instance-aware convolutions and eliminates the process of bounding boxes cropping and keypoint grouping.

\textcolor{black}{Overall, heatmap-based methods are more popular than the regression-based paradigms due to their higher accuracy. However, the heatmap computation process brings new open problems,  including expensive computational overhead and inevitable quantization error. }

\subsubsection{Video-Based Methods}
Human pose estimation on videos has also been a hot research topic. 
The video, by nature, brings more challenges such as \emph{camera shift, rapid object movement}, and \emph{defocus}, which result in frame quality deterioration frequently.
On the other hand, different from still images, there exist abundant temporal clues across video frames (\emph{e.g.}, \emph{temporal dependency} and \emph{geometric consistency}), which provide  valuable information for pose estimation.
\par
We observe that most existing methods are trained on static images. 
Directly applying the image-based models to videos (image sequence) might lead to unsatisfactory results since they fail to consider the temporal consistency across video frames. 
To conquer this dilemma, a large number of approaches have explored utilizing the additional temporal information to achieve higher pose detection accuracy. According to how the temporal information is exploited, we broadly divide these approaches into \emph{optical flow-based} \cite{Zhang_2015_ICCV, pfister2015flowing, song2017thin, zhang2018poseflow, chang2020towards}, \emph{RNN-based} (Recurrent Neural Networks) \cite{gkioxari2016chained, luo2018lstm, artacho2020unipose}, \emph{pose tracking-based} \cite{ yu2018multi, girdhar2018detect, wang2019ai, zhou2020temporal, wang2020combining, yang2021learning}, and \emph{key frame-based} \cite{charles2016personalizing, bertasius2019learning, nie2019dynamic, zhang2020key, liu2021deep} paradigms. 
Below, we elaborate these methods in detail.
\par
\textcolor{black}{\textbf{Optical Flow}\quad}
\textcolor{black}{\emph{Optical flow} models the apparent motion of individual pixels on the frame, attracting widespread attention \cite{Dosovitskiy_2015_ICCV, Ilg_2017_CVPR}.}
The optical flow across frames usually reveals the motions of the human subjects, which are obviously useful for pose estimation. 
\cite{pfister2015flowing} combines convolutional networks and optical flow into a uniform framework, which employs the flow field to align the features temporally across multiple frames, and utilizes the aligned features to improve the pose detection in individual frames.
\cite{song2017thin} presents a Thin-Slicing Network which computes the dense optical flow between every two frames to propagate the initial estimation of joint position through time, and uses a flow-based warping mechanism to align the joint heatmaps for subsequent spatiotemporal inference.
\textcolor{black}{
\cite{chang2020towards} focuses on human pose estimation in crowded scenes, which incorporates forward pose propagation and  backward pose propagation to refine the pose of the current frame. 
However, although the optical flow in these methods does contain useful features such as human motion information, the undesired background changes are also involved. 
The \emph{noisy} motion representation greatly hinders them from obtaining expected performance.}
\cite{zhang2018poseflow} proposes a novel deep motion representation, \emph{namely PoseFlow}, which is able to reveal human motion in videos while inhibiting some nuisance noises such as background and motion blur. The distilled robust flow representation can also be generalized to human action recognition tasks.

\textcolor{black}{The optical flow based representation can model the motion cues at the pixel level, which is favorable for capturing useful temporal information. However, the optical flow is only able to extract impure features and is quite sensitive to noises. }
\par
\textcolor{black}{\textbf{Recurrent Neural Network}\quad}
\textcolor{black}{Besides optical flow, \emph{Recurrent Neural Network (RNN)} also provides a way to model temporal contexts across frames. 
 RNN shows a promising performance in sequential prediction task, due to the nature that each output is jointly determined by the current input and the historical predictions.
Therefore, a group of approaches attempt to capture temporal contexts between video frames by RNN for improving pose estimation.}
\cite{gkioxari2016chained} presents a sequence-to-sequence model, which employs the chained convolutional networks to process input images, and combines historical hidden status and current images to predict current keypoint heatmaps.
\cite{luo2018lstm} extends the convolutional pose machine \cite{Wei_2016_CVPR} by using convolutional LSTM, which is able to model both spatial and temporal contexts for pose prediction.

\textcolor{black}{To our knowledge, existing RNN-based methods can effectively estimate human poses from the \emph{single-person} image sequence, yet they have not been applied to multi-person videos until now. We conjecture that RNN has difficulties in directly employing temporal information from multi-person videos, where extracting the temporal contexts of each person will be affected by the others.
} 
\par
\textcolor{black}{\textbf{Pose Tracking}\quad}
\textcolor{black}{To alleviate the issue of RNN, some methods that built upon the \emph{pose tracking} have been proposed, which establish a tracklet for each person in video frames to filter the interference of irrelevant information.}
\cite{girdhar2018detect} proposes a 3D Mask R-CNN (extension of Mask R-CNN \cite{He_2017_ICCV} to include a temporal dimension) to generate small clips for a single person, and leverages temporal information within the small clips to produce more accurate predictions. 
\cite{zhou2020temporal} proposes a pose estimation framework which consists of a temporal keypoint matching module and a temporal keypoint refinement module. Specifically, the temporal keypoint matching module gives reliable single-person pose sequences according to the keypoint similarities, and the temporal keypoint refinement module aggregates  poses within the sequence to correct original poses.
\cite{wang2020combining} designs a Clip Tracking Network and a Video Tracking Pipeline to establish the tracklet for each person, and extends the HRNet \cite{sun2019deep} to 3D-HRNet to perform temporal pose estimation for all tracklets.
\cite{yang2021learning} employs a graph neural network to learn the pose dynamics from the historical pose sequence, 
and incorporates the pose dynamics into the pose detection of  the current frame.

\textcolor{black}{Pose tracking-based methods show strong adaptation in the scene of multi-person. However, these models require computing feature similarity or pose similarity to create tracklets, which invokes an extra overhead for pose estimation.} 

\par
\textcolor{black}{\textbf{Key Frame Optimization}\quad}
In addition to exploiting temporal information from tracklets, it is also beneficial to select some key frames to refine the pose estimation of the current frame, what we refer to as \emph{keyframe-based} approaches.
\cite{charles2016personalizing} proposes a personalized video pose estimation framework, which leverages a few key frames with high-precision pose estimates to fine-tune the model.
\cite{bertasius2019learning} proposes a PoseWarper network which first warps poses of the labeled frames to the unlabeled (current) frame, and then aggregates all warped poses to predict the pose heatmaps of the current frame. 
\cite{zhang2020key} presents a keyframe proposal network to select the effective key frames, and proposes a learnable dictionary to reconstruct entire pose sequence from the selected key frames.
The work in \cite{liu2021deep} builds a dual consecutive framework for video pose estimation, termed DCPose\footnote{Link of DCPose Project: \url{https://github.com/Pose-Group/DCPose}}, which incorporates consecutive frames from dual temporal directions to improve the pose estimation in videos. Specifically, three modular components are designed. A Pose Temporal Merger encodes keypoint spatiotemporal context to generate effective searching scopes while a Pose Residual Fusion module computes weighted pose residuals in dual directions. These are then processed via a Pose Correction Network for efficient refining of pose estimations. \textcolor{black}{It is worthy mentioning that the DCPose \cite{liu2021deep} is able to fully leverage the temporal information from neighboring frames and achieves state-of-the-art performance on video-based human pose estimation.}

\subsubsection{Model Compression-Based Methods}
For practical applications on lightweight devices such as mobiles, a low-consumption and high-accuracy HPE method is urgently demanded. 
However, the majority of existing pose estimation models are oversized, which require extensive computational resources and fail to reach real-time computation. 
Consequently, these methods are usually low-efficient, which limits their potential usage especially for mobiles or wearable equipments. 
To alleviate this problem, many model compression based methods \cite{yu2021lite, li2021synthetic, zhang2019fastpose, nie2019dynamic, luo2018lstm} have been proposed to achieve the trade-off between accuracy and efficiency. These methods are able to significantly reduce model parameters with small accuracy decline.


\par
\cite{zhang2019fastpose} proposes a Fast Pose Distillation model that built upon the Teacher-Student network \cite{hinton2015distilling, zhou2018rocket, romero2014fitnets, wang2019deepvid, mirzadeh2020improved}, effectively transferring the human body structure knowledge from a strong teacher network (\emph{large model}) to a \emph{lightweight} student network.  Specifically, the $8$-stage Hourglass model is employed as the teacher network while a compact counterpart ($4$-stage Hourglass) is adopted as the student network. 
\textcolor{black}{\cite{luo2018lstm} proposes a lightweight LSTM architecture to perform video pose estimation. \cite{yu2021lite} proposes two schemes to reduce the parameters of HRNet: i) Simply applying the Shuffle-Block \cite{zhang2018shufflenet} to replace the basic block in \emph{vanilla} HRNet. ii) Designing a conditional channel weighting module, which learns the weights across multiple resolutions to replace the costly point-wise ($1 \times 1$) convolutions. By simplifying the original HRNet \cite{sun2019deep}, the Lite-HRNet \cite{yu2021lite} shows good performance with relatively fewer parameters.}

\subsubsection{Summary of Top-Down Framework}
The architecture of top-down framework comprises the following key components: an object detector for producing human bounding boxes, and a pose estimator for detecting human keypoint locations.
The object detector determines the performance of human proposal detection, and further influences pose estimation. 
The pose detector, on the other hand, is the core of the framework and directly determines the accuracy of pose estimation.
In summary, the top-down framework is highly scalable that can be constantly improved with advances of object detectors as well as pose detectors.
\subsection{Bottom-Up Framework}
The major discrepancy between bottom-up and top-down frameworks is whether the human detector is employed to detect the human bounding boxes. 
Compared to the top-down approaches, bottom-up approaches do not rely on human detection and directly perform keypoint estimation in the original image, thus reducing the computational overhead.
However, this procedure opens up a new challenge: How to judge the identities of estimated joints?
According to the way of determining the identities of estimated keypoints, we divide the bottom-up methods into \emph{human center regression-based} \cite{nie2018pose, geng2021bottom, nie2019single, nie2018human} , \emph{associate embedding-based} \cite{luo2021rethinking,  cheng2020higherhrnet, jin2019multi, newell2016associative}, and \emph{part field-based} \cite{hidalgo2019single, raaj2019efficient, kreiss2019pifpaf, Cao_2017_CVPR, jin2020differentiable, wang2020ktn, li2020simple, luo2018multi, kocabas2018multiposenet, pishchulin2016deepcut, insafutdinov2016deepercut, pishchulin2013poselet} approaches.

\textbf{Human Center Regression} The \emph{human center regression-based} approaches utilize a human center point to represent the person instance.
\cite{nie2019single} proposes a Single-stage multi-person Pose Machine that unifies person instance and body joint position representations. In \cite{nie2019single},  the root joints (center-biased points) are introduced to denote the person instances, and body joint locations are encoded into their displacements \emph{w.r.t.} the roots. 
\cite{geng2021bottom} predicts a human center map that indicates the person instance, and densely estimates a candidate pose at each pixel $q$ within the center map.
\par

\textbf{Associate Embedding} The \emph{associate embedding-based} approaches assign each keypoint an associate embedding, which is an instance representation for distinguishing different persons. 
\cite{newell2016associative} pioneers the embedding representation, where  each predicted keypoint has an additional  embedding vector that serves as a \emph{tag} to identify its human instance assignment.
\cite{jin2019multi} proposes a SpatialNet to detect body part heatmaps and  predict part-level data association in the input image. Specifically, the part-level data association is parameterized by the keypoint embedding.
\cite{cheng2020higherhrnet} follows the keypoints grouping in \cite{newell2016associative} and further proposes a Higher-Resolution Network to learn high-resolution feature pyramids, improving the pose estimation of small persons.
\cite{luo2021rethinking} focuses on the problems of  large variance of human scales and labeling ambiguities. This approach \cite{luo2021rethinking}  proposes a scale-adaptive heatmap regression model, which is able to adaptively adjust the standard deviation of the ground-truth gaussian kernels for each keypoint, and achieves high tolerance for different human scales and labeling ambiguities. 

\textbf{Part Field}
The \emph{part field-based} methods first detect keypoints and connections between them, and then perform keypoint grouping according to the keypoint connections.
\textcolor{black}{The representative work \cite{Cao_2017_CVPR} proposes a two-branch multi-stage CNN architecture, where one branch predicts the confident maps to denote the locations of keypoints and another branch  predicts the Part Affinity Fields to indicate the connective intensity between keypoints.
Then, \cite{Cao_2017_CVPR} applies a greedy algorithm to assemble different joints of the same person, according to the connective intensity between joints.
Inspired by \cite{Cao_2017_CVPR}, various attempts have been proposed.
\cite{kreiss2019pifpaf} utilizes a part intensity field to localize body parts, and employs a part association field to associate body parts with each other.
\cite{li2020simple} presents a novel keypoint associated representation of \emph{body part heatmaps} based on the Part Affinity Field \cite{Cao_2017_CVPR} for effective keypoint grouping.
}
Some approaches explore alternative representations of keypoint connection for keypoint grouping.
\cite{luo2018multi} proposes a multi-layer fractal network, which regresses the keypoint location heatmaps and infers kinships among adjacent joints to determine the optimal matched joint pairs.
\cite{jin2020differentiable} proposes a differentiable Hierarchical Graph Grouping network that converts the keypoint grouping into a graph grouping problem, and can be trained end-to-end with the keypoint detection network.

\textbf{Summary} Overall, the bottom-up approaches improve the efficiency of pose detection by eliminating the usage of additional object detection techniques.
Due to the high efficiency, the bottom-up methods are promising in practice applications. For example, the open source project\footnote{Link of OpenPose Project: \url{https://github.com/CMU-Perceptual-Computing-Lab/openpose}} of OpenPose \cite{cao2017realtime} has been extensively adopted in the industry.

\section{Network Training Refinement}\label{train}
From the perspective of the overall training pipeline in neural networks, the quantity and quality of data, training strategy, and loss function will impact the model performance.
According to the above key phases during training, we classify the network training refinement approaches into \emph{data augmentation techniques},  \emph{multi-task training strategies}, \emph{loss function constraints}, and \emph{domain adaption methods}.
Data augmentation techniques aim to increase the amount and diversity of the data.
Multi-task training strategies seek to capture informative features by sharing representations among related visual tasks.
Loss function constraints determine the optimization objective of the network. Domain adaption methods aim to help the network  adapt different datasets.
In this section, we introduce these methods in detail.
\subsection{Data Augmentation Techniques}
\textcolor{black}{Deep learning is typically data-driven, therefore data plays a crucial role in model training. A \emph{large-scale} and \emph{high-quality} dataset contributes to the robustness of models.
However, building such a wonderful dataset is time-consuming and expensive.}
To alleviate this problem, data augmentation techniques are adopted to increase the number and diversity of samples in datasets.

In 2D human pose estimation, common data augmentation techniques include random rotation, random scale, random truncation, horizontal flipping, random information dropping, and illumination variations.
Apart from  the above random schemes, several works \cite{peng2018jointly, moon2019posefix, huang2020aid, bin2020adversarial, zhou2017towards, wang2021human} have been studying  learnable data augmentation.
%
 \cite{peng2018jointly} proposes an enhancement network that generates difficult pose samples to compete against the pose estimator. 
 \cite{tang2019does} points out that state-of-the-art human pose estimation approaches have similar error distributions.
 \cite{moon2019posefix} generates synthetic poses based on the error statics in \cite{tang2019does} and employs the synthesized poses to train human pose estimation networks.
 \cite{bin2020adversarial} presents an adversarial semantic data augmentation using the generative  adversarial network (GAN \cite{goodfellow2014generative}), which enhances original images by pasting segmented body parts with different semantic granularities.
 \cite{wang2021human} introduces an AdvMix algorithm, in which a generator network confuses pose estimators by mixing various corrupted images, and a knowledge distillation network transfers clean pose structure knowledge to the target pose detector.
\subsection{Multi-Task Training Strategies}
Most of the human pose estimation models are designed for single-task learning. In this subsection, we focus on the multi-task learning models related to 2D human pose estimation.
\emph{Multi-task learning} aims at capturing informative features by sharing representations among related visual tasks.
Human parsing is a closely related task to human pose estimation, with the goal of segmenting the human body into semantic parts such as head, arms, and legs, etc.
Previous works \cite{ladicky2013human, dong2014towards, xia2017joint, nie2018human, liang2018look, duan2019trb} employ the human parsing information to improve the performance of 2D HPE.
\cite{xia2017joint} jointly solves the two tasks of human parsing and pose estimation, and utilizes the part-level segments to guide the keypoint localization. 
\cite{nie2018human} presents a parsing encoder and a pose model parameter adapter, which together learn to predict  parameters of the pose model to extract complementary features for human pose estimation.
\subsection{Loss Function Constraints}
Loss function determines the learning objective of the network, and greatly affects the performance of the model.
In this subsection, we summarize and discuss existing loss functions \cite{pishchulin2016deepcut, carreira2016human, he2017mask, sun2017compositional, ke2018multi, chen2018cascaded, li2020simple, yuan2020combined, zhou2020occlusion, luo2021rethinking} of 2D HPE. 
\par
The standard and common loss function of human pose estimation is the $L_2$ distance. Training aims to minimize the total L2 distance between prediction and ground truth heatmaps for all joints. The cost function is defined as:
	\begin{align}
		L = \frac{1}{N} * \sum_{j=1}^N v_j \times ||G\left(j\right) - P\left(j\right)||^2
	\end{align}
Where $G(j)$, $P(j)$ and $v_j$ respectively denote the ground truth heatmap, prediction heatmap and visibility for joint $j$.
The symbol $N$ denotes the number of joints.
\par
\cite{ke2018multi} presents a multi-scale human structure-aware loss which captures the structural information of the human body.
The \emph{structure-aware loss} at the $i^{th}$ feature scale can be expressed as follows:
\begin{align}
	L^i = \frac{1}{N}\sum_{j=1}^N||P_j^i - G_j^i||_2 + \alpha \sum_{i=1}^N||P_{S_j}^i - G_{S_j}^i||_2,
\end{align}
where $P_j$ and $G_j$ denote the predicted and labeled $j^{th}$ keypoint heatmaps, $P_{S_j}$ and  $G_{S_j}$ are the group of the heatmaps from  keypoint $j$ and its neighbors, respectively.
\par
\cite{chen2018cascaded} proposes an online hard keypoints mining, which first computes the regular $L_2$ loss for all keypoints, and then additionally punishes top-$M$ hard keypoints.
%
This loss function increases the penalty of the difficult keypoints, and improves the network performance.
\par
\cite{yuan2020combined} presents a combined distillation loss for the HRNet, which consists of a structure loss (STLoss), a pairwise inhibition loss (PairLoss), and a probability distribution loss (PDLoss).
Specifically, the STLoss enforces the network to learn human structures at earlier phase to combat against pose occlusions, and the PairLoss alleviates the problem of similar joint misclassification especially in crowded scenarios.
The PDLoss guides the learning of the distribution of final heatmaps.
\par
\par
\subsection{Domain Adaption Methods}
Human pose estimation has been widely investigated with much focus on supervised learning that requires sufficient pose annotations. However, in real applications, pretrained pose estimation models usually need be adapted to a new domain with no labels or sparse labels. 
Therefore, several domain adaptation methods  \cite{guo2018multi, hidalgo2019single, xu2020alleviating, li2021synthetic} leverage a labeled source domain to learn a model that performs well on an unlabeled or
sparse labeled target domain.
%
%
\par

\cite{xu2020alleviating} proposes a domain adaptation method for 2D HPE, which accomplishes both the human body-level topological structure alignment and fine-grained feature alignment in different datasets.
\cite{guo2018multi} proposes a multi-domain pose network that is able to train the model on multiple dataset simultaneously, which obtains a better pose representation in a multi-domain learning fashion.
\textcolor{black}{\cite{li2021synthetic} proposes an online coarse-to-fine pseudo label updating strategy to reduce the gap between the synthetic and real data, which have demonstrated strong generalization ability for animal pose estimation.
\cite{li2021synthetic} is able to softens the label noises and thereby delivers state-of-the-art results on multiple animal benchmark datasets.}
\section{Post Processing Approaches}\label{post}
\textcolor{black}{Instead of predicting the final keypoint locations at once, some approaches first estimate an initial pose and then optimize it with some post-processing operations, which we refer to as \emph{post processing} methods. 
We divide these methods into two categories, \emph{i.e.}, quantization error and pose resampling.}
For the heatmap representation of keypoints, the conversion from heatmap to coordinate space inevitably occurs errors, which leads to \emph{quantization errors}.
Suppressing such quantization errors will boost the performance of numerous heatmap-based models.
\textcolor{black}{On the other hand, an out-of-the-box pose refinement technique, \emph{pose resampling}, aims at resampling favorable pose representations to improve the initial estimations.}
In what follows, we elaborate on the above approaches.
\subsection{Quantization Error}
\par
\textcolor{black}{The extensively adopted heatmap based pose representation requires decoding the 2D coordinates $(x, y)$ of joints from estimated keypoint heatmaps.
In particular, we take the position of the maximum activation value from the predicted heatmap as the keypoint coordinates.
However, the predicted gaussian heatmaps do not always conform to the  standard gaussian distribution and potentially contain multiple peak values, which degrades the accuracy of the coordinate computation. 
To address the issue, \cite{zhang2020distribution} proposes a distribution-aware architecture that first performs heatmap distribution modulation to adjust the shape of predicted heatmaps and then employs a new coordinate decoding method to accurately obtain the final keypoint locations. This approach reduces mistakes of the conversion from heatmaps to coordinates, and improves the performance of existing heatmap-based models.}
\cite{huang2020devil} quantitatively analyzes the common biased data processing on 2D HPE, and further processes data based on unit length instead of pixel, which obtains aligned pose results when flipping is performed in inference.
\textcolor{black}{Furthermore, this approach introduces an encoding-decoding method, which is theoretically error-free for the transformation of keypoint locations between heatmaps and coordinates.}

\textcolor{black}{
On the other hand,  the non-differentiable property of the maximum operation in the decoding process also introduces quantization errors.
To address this problem, a group of researches  \cite{luvizon2019human, sun2018integral}  attempt to design differentiable algorithms.
\cite{luvizon2019human} proposes  a fully differentiable and end-to-end trainable regression approach, which utilizes the novel Soft-argmax function to convert feature maps directly to keypoint coordinates.
\cite{sun2018integral} proposes an integral method to tackle the problem of non-differentiable from heatmaps to coordinates.}

\subsection{Pose Resampling}
\textcolor{black}{A wide spectrum of pose estimators \cite{sun2019deep, xiao2018simple} directly take the model output as final estimates.
However, these estimations can be further improved by a model-agnostic pose resampling technique.
A line of work considers fine-tuning of the initial estimation with additional pose cues.
\cite{moon2019posefix} proposes a model-agnostic PoseFix method that estimates a refined pose from a tuple of an input image and an input pose, where the input pose is derived from the estimations of existing methods.}
\cite{qiu2020peeking} proposes to first localize the visible joints based on visual information by an existing pose estimator, and then estimate the invisible joints by an Image-Guided Progressive GCN module that combines image context and pose structure cues.
\cite{wang2020graph} proposes a two-stage and model-agnostic framework, namely Graph-PCNN, which employs an existing pose estimator for coarse keypoint localization, and designs a graph pose refinement module to produce more accurate localization results.
\par
\textcolor{black}{The above pose resampling methods are designed for static images, and 
some approaches explore the pose resampling techniques for videos.
Specifically, these methods \cite{wang2020combining, liu2021deep, yang2021learning, bertasius2019learning, zhou2020temporal} perform pose aggregation to integrate multiple estimated poses of current frame to refine estimations.
Normalization is commonly leveraged to aggregate multiple pose predictions \cite{liu2021deep, bertasius2019learning, yang2021learning}, where the various predictions are treated equally.}
\cite{wang2020combining} introduces the Dijkstra algorithm \cite{dijkstra1959note} to solve the problem of optimal keypoint locations, which first employs the mean shift algorithm \cite{comaniciu2002mean} to group all pose hypotheses into various clusters, and subsequently selects the keypoint with closest distance to the cluster center as the optimal result.
\cite{zhou2020temporal} utilizes the pose similarity between the neighboring frames and the current frame to biasedly aggregate features, and then employs a convolutional neural network to decode current heatmaps from the aggregated features.
\section{Datasets and Evaluation}\label{dataset}
Benchmark datasets form the basis of deep learning models, and also provide a common foundation for measuring and comparing the performance of competing approaches.
In this section, we present the major \emph{benchmark datasets}, \emph{evaluation metrics}, and \emph{performance comparisons} for human pose estimation.
\renewcommand\arraystretch{1.7}
\begin{table*}
\vspace{0.2em}\caption{A summary of 2D human pose estimation benchmark datasets. \emph{Upper Poses}, \emph{Full Poses}, \emph{Various Poses} denotes the upper body poses, singular full body poses and various body poses, respectively.}\label{tab2:dataset}
  \resizebox{1\textwidth}{!}{
  \begin{tabular}{l|c|c|c|c|c|c|c|c|ccc}
    \hline
    \multirow{2}{*}{Dataset Name} & \multirow{2}{*}{Year} & \multirow{2}{*}{Single-Person} & \multirow{2}{*}{Multi-Person} 
    & \multirow{2}{*}{Upper Poses} & \multirow{2}{*}{Full Poses} & \multirow{2}{*}{Various Poses} 
    & \multirow{2}{*}{Number of Joints} &\multirow{2}{*}{Evaluation Metric}
	 & \multicolumn{3}{c}{Number of Images / Videos} \\ \cline{10-12}       
	 & & & & & & & & &Train &Val &Test \cr
	 \hline
	 \multicolumn{12}{c}{\textbf{Image}-Based Datasets for Human Pose Estimation.}\cr
	 \hline
	 LSP \cite{johnson2010clustered} 			&2010	&\checkmark	& & &\checkmark &	 & 14	& PCP &$1,000$   &- &$1,000$ \cr
	 LSP-Extended \cite{johnson2011learning} 	&2011	&\checkmark	& & &\checkmark &	 & 14	& PCP &$10,000$ &- &- \cr
	 \hline
	 Flic \cite{sapp2013modec} 		   &2013	&\checkmark	& &\checkmark & &	 & 10	& PCP &$5,000$   &- &$1,016$ \cr
	 Flic-Full \cite{sapp2013modec}    &2013	&\checkmark	& &\checkmark & &	 & 10	& PCP &$20,928$ &- &- \cr
	 Flic-Plus \cite{tompson2014joint} &2013	&\checkmark	& &\checkmark & &	 & 10	& PCP &$17,380$ &- &- \cr
	\hline 
	 MPII \cite{andriluka20142d} 	&2014	&\checkmark	&		    & & &\checkmark	 & 16	& PCPm/PCKh &$28,821$ &- &$11,701$ \cr
	 MPII \cite{andriluka20142d} 	&2014	&			&\checkmark & & &\checkmark	 & 16	& PCKh       &$3,800$  &- &$1,700$ \cr
	\hline
	 COCO \cite{lin2014microsoft} 	&2017	&	&\checkmark & & &\checkmark	 & 17	& AP &$57,000$ &$5,000$ &$20,000$ \cr
	\hline
	 AIC-HKD \cite{wu2017ai} 			&2017	&	&\checkmark & & &\checkmark	 & 14	& mAP &$210,000$ &$30,000$ &$60,000$\cr
	 \hline
	 CrowdedPose \cite{li2019crowdpose} &2019	&	&\checkmark & & &\checkmark	 & 14	& mAP &$10,000$ &$2,000$ &$8,000$ \cr
	\hline
	\multicolumn{12}{c}{\textbf{Video}-Based Datasets for Human Pose Estimation.}\cr
	\hline
	Penn Action \cite{zhang2013actemes} 	&2013	&\checkmark	& & &\checkmark &	 & 13	& mAP &$1,000$ &- &$1,000$ \cr
	\hline
	JHMDB \cite{jhuang2013towards} 	    &2013	&\checkmark	& & &\checkmark &	 & 15	& mAP &600 &- &300 \cr
	\hline
	PoseTrack2017 \cite{iqbal2017pose} 			&2017	&	&\checkmark & & &\checkmark	 & 15	& mAP &250 &50 &214 \cr
	PoseTrack2018 \cite{andriluka2018posetrack} &2018	&	&\checkmark & & &\checkmark	 & 15	& mAP &593 &170 &375 \cr
	\hline
	HiEve \cite{lin2020human} 					&2020   &	&\checkmark & & &\checkmark	 & 14	& mAP &19 &- &13 \cr
	\hline
    \end{tabular}}
\end{table*}
\begin{figure*}[t]
\begin{center}
\includegraphics[width=1\linewidth]{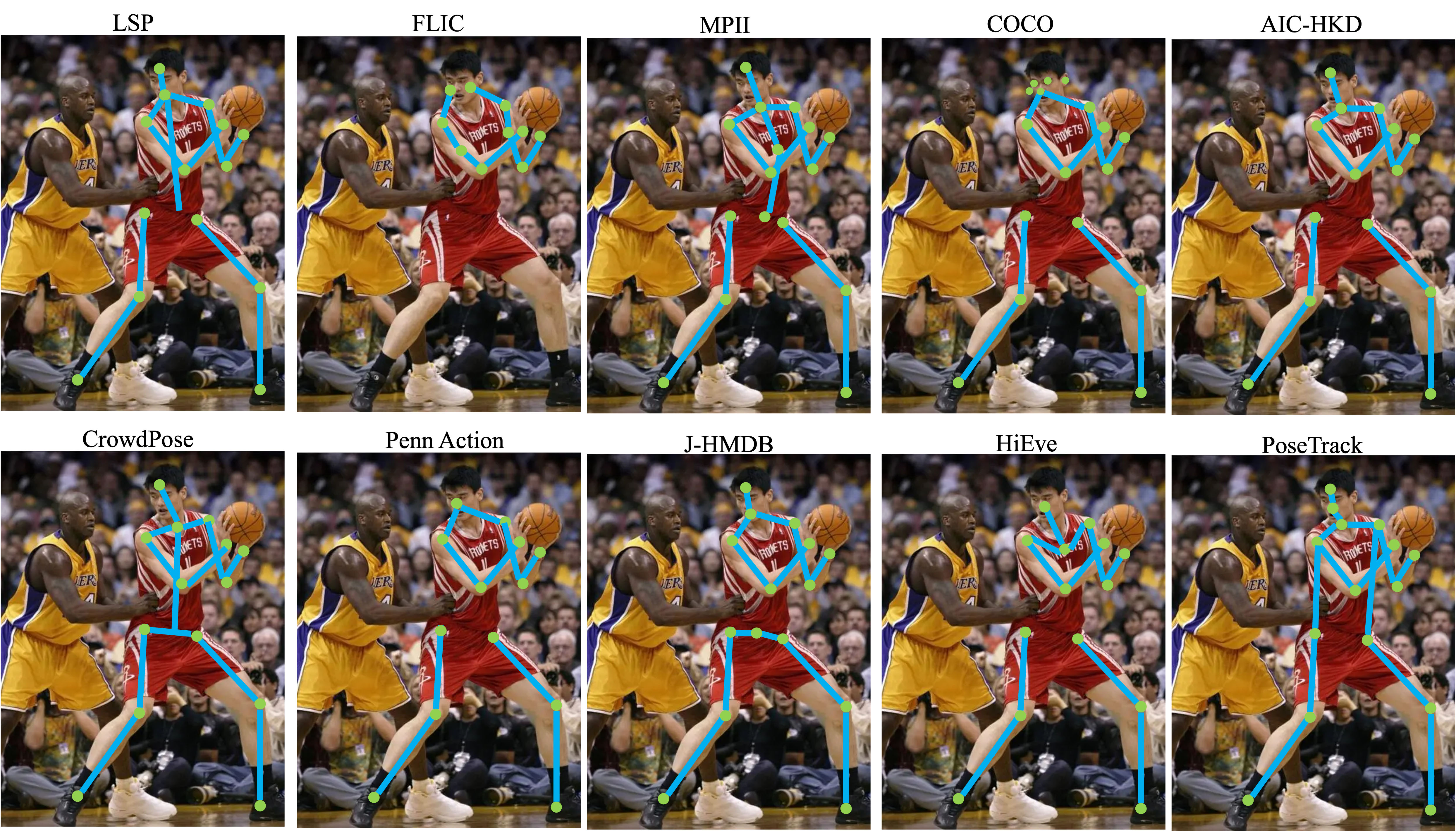}
\end{center}
\caption{Illustration of pose annotations for different benchmark datasets including LSP, FLIC, MPII, COCO, AIC-HKD, CrowdedPose, Penn Action, J-HMDB, HiEve, and PoseTrack.}
\label{fig:label}
\end{figure*}
\renewcommand\arraystretch{1.7}
\begin{table*}
\vspace{0.2em}\caption{Performance comparisons of state-of-the-art methods including \emph{top-down} approaches, \emph{bottom-up} approaches and \emph{small networks} on \textbf{COCO} benchmark dataset (test-dev2017).}\label{tab:img}
  \resizebox{1\textwidth}{!}{
  \begin{tabular}{l|c|c|c|c|ccccc|ccc}
    \hline
 Method &Backbone &Input size &Parameters   &GFLOPs   &AP &AP$^{50}$ &AP$^{75}$ &AP$^{M}$ &AP$^{L}$ &AR &AR$^{M}$ &AR$^{L}$\cr
	\hline
	\multicolumn{11}{c}{Top-down framework: human detection and individual keypoint detection.}\cr
    \hline
    Mask-RCNN \cite{he2017mask}   	    		&ResNet-50   &-                  &-       &-      &63.1    &87.3  &68.7 &57.8 &71.4  &- &-&-\cr
    G-RMI \cite{papandreou2017towards}  		&ResNet-101  &$353 \times 257$   &42.6M   &57.0   &64.9    &85.5  &71.3 &62.3 &70.0  &69.7&-&-\cr
Integral Pose \cite{sun2018integral}			&ResNet-101  &$256 \times 256$   &45.0M   &11.0   &67.8    &88.2  &74.8 &63.9 	&74.0  &-&-&-\cr
G-RMI$+$extra data \cite{papandreou2017towards}	&ResNet-101  &$353 \times 257$   &42.6M   &57.0   &68.5    &87.1  &75.5 &65.8 	&73.3  &73.3&-&-\cr
CPN \cite{chen2018cascaded}   	    	&Resnet-Inception	 &$384 \times 288$   &-       &-      &72.1    &91.4  &80.0 &68.7 	&77.2  &78.5&-&-\cr
RMPE \cite{fang2017rmpe}   	    		&Stacked Hourglass	 &$320 \times 256$   &28.1M   &26.7   &72.3    &89.2  &79.1 &68.0 	&78.6  &-&-&-\cr
CFN \cite{huang2017coarse}   	    			& -			 &-                  &-       &-      &72.6    &86.1  &69.7 &78.3 	&64.1  &-&-&-\cr
CPN (ensemble) \cite{chen2018cascaded}  &Resnet-Inception	 &$384 \times 288$   &-       &-      &73.0    &91.7  &80.9 &69.5 	&78.1  &79.0&-&-\cr
SimpleBaseline \cite{xiao2018simple}			&ResNet-152	 &$384 \times 288$   &68.6M   &35.6   &73.7    &91.9  &81.1 &70.3 	&80.0  &79.0&-&-\cr
HRNet-W32 \cite{sun2019deep}					&HRNet-W32	 &$384 \times 288$   &28.5M   &16.0   &74.9    &92.5  &82.8 &71.3 	&80.9  &80.1&-&-\cr   
HRNet-W48 \cite{sun2019deep}					&HRNet-W48	 &$384 \times 288$   &63.6M   &32.9   &75.5    &92.5  &83.3 &71.9 	&81.5  &80.5&-&-\cr  
DARK \cite{zhang2020distribution}				&HRNet-W48	 &$384 \times 288$   &63.6M   &32.9   &76.2    &92.5  &83.6 &72.5 	&82.4  &81.1&-&-\cr 
UDP  \cite{huang2020devil}						&HRNet-W48	 &$384 \times 288$   &63.6M   &33.0   &\textbf{76.5}    &92.7  &84.0 &73.0 	&82.4  &81.6&-&-\cr 
HRNet-W48$+$extra data \cite{sun2019deep}		&HRNet-W48	 &$384 \times 288$   &63.6M   &32.9   &77.0    &92.7  &84.5 &73.4 	&83.1  &82.0&-&-\cr  
DARK$+$extra data \cite{zhang2020distribution}	&HRNet-W48	 &$384 \times 288$   &63.6M   &32.9   &\textbf{77.4}    &92.6  &84.6 &73.6 	&83.7  &82.3&-&-\cr   
\hline
	\multicolumn{11}{c}{Bottom-up framework: keypoint detection and grouping.}\cr
    \hline
 AE \cite{newell2016associative}			&-	 &512   &-   &-   &63.0    &85.7  &68.9 &58.0 	&70.4  &- &- &-\cr  
 AE+refinement \cite{newell2016associative}	&-	 &512   &-   &-   &65.5    &86.8  &72.3 &60.6 	&72.6  &70.2 &64.6 &78.1\cr     
DirectPose \cite{tian2019directpose}		&-	 &800   &-   &-   &64.8    &87.8  &71.1 &60.4 	&71.5  &- &- &-\cr 
SimplePose \cite{li2020simple}				&-	 &512   &-   &-   &68.1    &-     &-    &66.8 	&70.5  &72.1 &- &-\cr  
HGG \cite{jin2020differentiable}			&-	 &512   &-   &-   &67.6    &85.1  &73.7 &62.7 	&74.6  &71.3 &- &-\cr  
PersonLab \cite{papandreou2018personlab}	&-	 &1401  &-   &-   &68.7    &89.0  &75.4 &64.1 	&75.5  &75.4 &69.7 &83.0\cr  
Point-set Anchors \cite{wei2020point}		&-	 &640  	&-   &-   &68.7    &89.9  &76.3 &64.8 	&75.3  &74.8 &69.6 &82.1\cr 
HrHRNet-W48$+$AE \cite{cheng2020higherhrnet}&HRNet-W48	 &640  	&-   &-   &70.5    &89.3  &77.2 &66.6 	&75.8  &- &- &-\cr  
DEKR-W48 \cite{geng2021bottom}		&HRNet-W48	 &640  	&-   &-   &71.0    &89.2  &78.0 &67.1 	&76.9  &76.7 &71.5 &83.9\cr 
SWAHR$+$HrHRNet-W48 \cite{luo2021rethinking}&HRNet-W48	 &-  	&-   &-   &\textbf{72.0}    &90.7  &78.8 &67.8 	&77.7  &- &- &-\cr  
\hline
	\multicolumn{11}{c}{Small networks}\cr
	\hline
Small HRNet \cite{yu2021lite}   &HRNet-W16 	 &$384 \times 288$   &1.3M  &1.21  &55.2  &85.8   &61.4  &51.7 &61.2 &61.5  &-&-\cr
MobileNetV2 $1\times$ \cite{sandler2018mobilenetv2} &MobileNetV2 &$384 \times 288$ &9.8M &3.33  &66.8  &90.0   &74.0  &62.6 &73.3 	&72.3  &-&-\cr
ShuffleNetV2 $1\times$ \cite{ma2018shufflenet}   &ShuffleNetV2 &$384 \times 288$   &7.6M  &2.87  &62.9  &88.5   &69.4  &58.9 &69.3 	&68.9  &-&-\cr	
Lite-HRNet \cite{yu2021lite}  &Lite-HRNet-30 &$384 \times 288$   &1.8M  &0.70  &\textbf{69.7}  &90.7   &77.5  &66.9 &75.0 	&75.4  &-&-\cr
    \hline 
    \end{tabular}}
\end{table*}

\renewcommand\arraystretch{1.7}
\begin{table*}
\vspace{0.2em}\caption{Performance comparisons of state-of-the-art methods on \textbf{PoseTrack2017} benchmark dataset (validation and test sets). Pretrain denotes the backbone model has been pretrained on COCO keypoint detection dataset.}\label{tab:video}
  \resizebox{1\textwidth}{!}{
  \begin{tabular}{l|c|c|c|c|c|c|c|c|c|c|c}
    \hline
     Method 		&Backbone &Pretrain &Additional Training Data   &Head   &Shoulder &Elbow &Wrist   &Hip    &Knee   &Ankle   &{\bf Mean}\cr
	\hline
	\multicolumn{12}{c}{Dataset: \textbf{PoseTrack2017 Validation} set.}\cr
    \hline
    PoseTracker\cite{girdhar2018detect}     &ResNet-3D &Y &COCO &$67.5$ &$70.2$   &$62.0$      &$51.7$  &$60.7$ &$58.7$ &$49.8$  &{$60.6$}\cr
     PoseFlow\cite{xiu2018pose}         	& - & - & MPII Pose + COCO	&$66.7$ & $73.3$  &$68.3$      &$61.1$  &$67.5$ &$67.0$ &$61.3$  &{$66.5$}\cr
JointFlow\cite{doering2018joint}        	& - & - & -		& -     & -       &-           &-       &-      &-      &-       &{$69.3$}\cr
     FastPose\cite{zhang2019fastpose}   	& - & - & -		&$80.0$ &$80.3$   &$69.5$      &$59.1$  &$71.4$ &$67.5$ &$59.4$  &{$70.3$}\cr
SimpleBaseline\cite{xiao2018simple}     	&ResNet-50  &N &COCO &$79.1$ &$80.5$   &$75.5$      &$66.0$  &$70.8$ &$70.0$ &$61.7$  &{$72.4$}\cr
SimpleBaseline\cite{xiao2018simple}     	&ResNet-152 &N &COCO	&$81.7$ &$83.4$   &$80.0$      &$72.4$  &$75.3$ &$74.8$ &$67.1$  &{$76.7$}\cr
  STEmbedding\cite{jin2019multi}    &$4$-stage Stacked Hourglass &Y  &-	&$83.8$ &$81.6$   &$77.1$      &$70.0$  &$77.4$ &$74.5$ &$70.8$  &{$77.0$}\cr
        HRNet\cite{sun2019deep}         	&HRNet-W48 &Y &COCO	 &$82.1$ &$83.6$   &$80.4$      &$73.3$  &$75.5$ &$75.3$ &$68.5$  &{$77.3$}\cr
         MDPN\cite{guo2018multi}        	&SimpleBaseline &Y &MPII Pose + COCO		&$85.2$ &$88.5$   &$83.9$      &$77.5$  & $79.0$&$77.0$ &$71.4$  &{$80.7$}\cr
 Dynamic\cite{yang2021learning} 		&HRNet-W48 &Y &COCO					&$88.4$ &$88.4$   &$82.0$      &$ 74.5$ &$79.1$ &$78.3$ &$73.1$  &{ $81.1$}\cr
 PoseWarper\cite{bertasius2019learning} 	&HRNet-W48 &Y &COCO					&$81.4$ &$88.3$   &$83.9$      &$ 78.0$ &$82.4$ &$80.5$ &$73.6$  &{ $81.2$}\cr
    \bf DCPose\cite{liu2021deep}  &HRNet-W48 &Y &COCO	&$\bf 88.0$  &$\bf 88.7$ &$\bf 84.1$ &$\bf 78.4$ &$\bf 83.0$ &$\bf 81.4$ &$\bf 74.2$ &$\bf 82.8$\cr
    \hline
    \multicolumn{12}{c}{Dataset: \textbf{PoseTrack2017 Test} set ( Results from the PoseTrack official leaderboard).}\cr
    \hline
    PoseTracker\cite{girdhar2018detect} &ResNet-3D &Y &COCO 				& -     & -       &-           &$51.5$  &-       &-        &$50.17$ &{$59.6$}\cr
    PoseFlow\cite{xiu2018pose}     	& - & - & MPII Pose + COCO        		&$64.9$  &$67.5$&$65.0$ &$59.0$ &$62.5$ &$62.8$  &$57.9$   &{$63.0$}\cr
	JointFlow\cite{doering2018joint}	& - & - & -	       					&-       &-     &-      &$53.1$ &-      &-       &$50.4$   &{$63.4$}\cr
	KeyTrack\cite{snower202015}    	& - & - &COCO        				    &-       &-     &-      &$71.9$ &-      &-       &$65.0$   &{$74.0$}\cr
	DetTrack\cite{wang2020combining}&3D-HRNet &Y &COCO   					&-       &-     &-      &$69.8$ &-      &-       &$65.9$   &{$74.1$}\cr
	SimpleBaseline\cite{xiao2018simple}&ResNet-152 &N &COCO    				&$80.1$ &$80.2$ &$76.9$ &$71.5$ &$72.5$ &$72.4$  &$65.7$   &{$74.6$}\cr
	HRNet\cite{sun2019deep}   	&HRNet-W48 &Y &COCO              			&$80.1$ &$80.2$ &$76.9$ &$72.0$ &$73.4$ &$72.5$  &$67.0$   &{$74.9$}\cr
	PoseWarper\cite{bertasius2019learning} 		&HRNet-W48 &Y &COCO			&$79.5$ &$84.3$ &$80.1$ &$75.8$ &$77.6$ &$76.8$  &$70.8$   &$77.9$\cr
    \bf DCPose\cite{liu2021deep}  &HRNet-W48 &Y &COCO	&$\bf 84.3$&$\bf 84.9$&$\bf 80.5$&$\bf 76.1$&$\bf 77.9$&$ \bf 77.1$&$\bf 71.2$&$\bf 79.2$\cr
    \hline
    \end{tabular}}
\end{table*}
\subsection{Benchmark Datasets}
Prior to the flourishing of deep learning, there are plenty of human pose datasets for specific task scenarios, including upper body pose datasets \cite{marin2014detecting, eichner2009better, everingham2010pascal, eichner2010we, sapp2011parsing, eichner2012human} and full-body pose dataset \cite{wang2011learning, li2007and, andriluka20142d, gong2016human}.
In this section, we investigate the datasets that are commonly used for deep learning, as summarized in Table \ref{tab2:dataset}. The corresponding pose annotations are depicted in Fig. \ref{fig:label}.
\par
\textbf{Leeds Sports Pose (LSP) Dataset}\quad
The LSP dataset contains a total number of $2,000$ images of full body poses (including 14 joints), $1,000$ images for training and test, respectively. 
This database is collected from the images tagged \emph{athletics, badminton, baseball, gymnastics, parkour, soccer, tennis, and volleyball} in the Flickr\footnote{Link of Flickr: \url{https://www.flickr.com/}}.
The LSP dataset is subsequently extended to the LSP-Extended dataset which contains over $10,000$ training images.
Datasets have been publicly available at \url{https://sam.johnson.io/research/lsp.html}.
\par
\textbf{Frames Labeled in Cinema (FLIC) Dataset}\quad
The FL-IC dataset consists of about $5,000$ images drawn from popular Hollywood movies, with $4,000$ images for training and $1,000$ images for test.
During labeling the keypoints, an object detector \cite{bourdev2009poselets} is first leveraged on the Flic dataset to give the  human candidates (roughly $20,000$ examples). These are then sent to the crowdsourcing marketplace \emph{Amazon Mechanical Turk} to obtain the ground truth poses including $10$ upper body joints. Severely occluded or non-frontal persons are manually cleaned to form the Flic-Full dataset.
These datasets have been publicly available at \url{https://bensapp.github.io/flic-dataset.html}.
\par
\textbf{MPII Human Pose Dataset}\quad
The MPII dataset contains $28,821$ images for training and $11,701$ images for test. 
This dataset covers various human activities including recreational, occupational,  house holding activities, and involves over $40,000$ individual persons under a wide spectrum of viewpoints.
The pose annotations include 15 human joints and occlusion labels.
This dataset has been publicly available at \url{http://human-pose.mpi-inf.mpg.de/}.
\par
\textbf{Common Objects in Context (COCO) Dataset}\quad
Microsoft COCO dataset is one of the most commonly used large-scale vision benchmark datasets, containing a total number of $330,000$ images with over $200,000$ annotated images for vision tasks such as object detection, segmentation, captioning, superpixel stuff segmentation and pose estimation, \emph{etc}.
For 2D human pose estimation, 200,000  labeled images with 250,000 pose annotations are included.
Pose annotations with 17 joints on training and validation sets are publicly available, and labels of test set are unavailable.
The COCO dataset has become the most popular benchmark in image-based human pose estimation. Therefore, we subsequently report performance comparisons among different algorithms in this dataset.
The COCO dataset for 2D human pose estimation can be obtained in \url{https://cocodataset.org/#keypoints-2020}.
\par
\textbf{AI Challenger (AIC) Dataset}\quad
The AIC dataset consists of three sub-datasets: human keypoint detection (HKD), large-scale attribute dataset and image Chinese captioning, respectively.
HKD contains $300,000$ images with a total of $700,000$ human instances labeled by 14 keypoints.
These images are collected from the Internet search engine with an emphasis on daily activates for ordinary people.
The link of official website is: \url{https://challenger.ai/}.
\par
\textbf{CrowdedPose Dataset}\quad
The CrowdedPose dataset is designed for the crowded scenarios, which contains $20,000$ images about $80,000$ individual persons. This dataset has a  split ratio of $5:1:4$ for training, validation, and test sets.
The dataset is collected by randomly sampling 30,000 images from three public benchmarks according to the \emph{Crowd Index} (a measurement of crowding level for a given image).
This dataset is available at \url{https://github.com/Jeff-sjtu/CrowdPose}.
\par
\textbf{Penn Action Dataset}\quad
The Penn Action dataset is an unconstrained human action dataset, which contains $2,326$ video clips derived from YouTuBe, and covers 15 type of actions. There are $1,258$ videos for training and $1,068$ videos for test.
Each person in images is labeled with 13 keypoints, and both joint coordinates and visibility are provided.
This dataset is available at \url{http://dreamdragon.github.io/PennAction/}.
\par
\textbf{Joint-Annotated Human Motion DataBase (JHMDB) Dataset}\quad
JHMDB dataset is a fully annotated dataset for human action recognition and human pose estimation, which contains 21 action categories including \emph{bru-sh hair, catch, clap, climb stairs,} and so on.
A subset of JHMDB that involves all visible joints, termed sub-JHMDB, are used for video-based 2D HPE. 
This subset contains 316 video clips with 12 action categories, and each person is annotated with 15 joints.
These datasets are available at \url{http://jhmdb.is.tue.mpg.de/}.
\par
\textbf{PoseTrack Dataset}\quad
PoseTrack is a large-scale public dataset for human pose estimation and articulated tracking, which  includes challenging situations with complicated movement of highly occluded people in crowded environments. The PoseTrack2017 dataset contains 514 video clips with $16,219$ pose annotations, and the PoseTrack2018 dataset greatly increased the number of video clips to $1,138$ with a total of $153,615$ pose annotations.
In training videos, dense annotations for 30 center frames
of a video are provided. In validation videos, human poses are annotated every four frames. Both datasets label 15 joints, with an additional annotation label for joint visibility.
These datasets are available at \url{https://posetrack.net}.
\par
\textbf{Human-Centric Video Analysis in Complex Events (HiEve) Dataset}\quad
HiEve is the largest dataset for video-based human pose estimation, which contains 31 videos with a total of  $1,099,357$ annotated poses, and labels 14 keypoints.
The HiEve dataset incorporates three human-centered understanding tasks, including human pose estimation, pose tracking, and action recognition.
The HiEve dataset is publicly available at \url{http://humaninevents.org/}.
\subsection{Evaluation Metrics}
Accuracy is the fundamental measurement of performance comparisons between different methods.
In Table \ref{tab2:dataset}, we list the metrics used to compute the accuracy of models in different datasets.
In what follows, we focus on the evaluation metrics of model accuracy.
\par
\textbf{Percentage of Correctly Estimated Body Parts (PCP)}\quad
The PCP metric reflects the accuracy of localized body parts. An estimated part is considered correct if its endpoints lie within a threshold, which can be a fraction of the length of the ground truth segment at its annotated location \cite{eichner20122d}.
In addition to the mean PCP of all body parts, separate body limbs PCP such as torso, upper legs and head are also usually reported.
Similar to the PCP metric, PCPm utilizes $50\%$ of the mean ground-truth segment length over the entire test as the matching threshold \cite{andriluka20142d}.
\par
\textbf{Percentage of Correct Keypoints (PCK)}\quad
PCK \cite{yang2012articulated} measures  the accuracy of the localized body keypoints, and a candidate joint is considered correct if it lies within a matching threshold.
The threshold for matching of the keypoint position to the ground-truth can be defined as a fraction of the human bounding box size (denoted as \emph{PCK}), and $50\%$ of the head segment length (denoted as \emph{PCKh}).
\par
\textbf{Average Precision (AP)}\quad
The AP metric is defined on the basis of the Object Keypoint Similarity (OKS) \cite{lin2014microsoft} that evaluates the similarity between predicted and ground-truth keypoints.
The Average Precision score under different OKS thresholds \emph{N} is denoted as AP@\emph{N}.
For the image-based human pose estimation, mean average precision (\textbf{mAP}) is the mean value of AP scores at all OKS  thresholds.
In video-based human pose estimation,  mAP averages the AP scores of each joint.
\subsection{Performance Comparisons}
In order to comprehensively provide a performance comparison for different human pose estimation algorithms, we pick two representative benchmark datasets: \emph{COCO} and \emph{PoseTrack2017}.
The performance of image-level human pose estimation models on COCO dataset are presented in Table \ref{tab:img}.
HRNet-W48 is a powerful backbone network with excellent performance for keypoint localiztion, and UDP network that builds upon the HRNet achieves state-of-the-art results without extra training data.
The bottom-up approaches remain a wide gap (5.4 mAP) compared to the top-down approaches.
In additional to the model accuracy, the efficiency is also important especially for practical applications.
To this end, we report some approaches that aim at designing small networks, as summarized in Table \ref{tab:img}.
Lite-HRNet achieves a better trade-off between accuracy and speed, which obtains the accuracy of $69.7$ mAP with parameters of $1.8$ M.
\par
We also report the performance of various video-based models on PoseTrack2017 dataset in Table \ref{tab:video}.
The DCPose employs abundant temporal information from adjacent frame to facilitate the current pose estimation, consistently establishing new state-of-the-arts on both validation and test sets.
\section{Discussion}\label{discussion}
In this section, we first discuss the open questions of the current 2D human pose estimation, including model generalization and datasets.
Subsequently, we introduce the incompletely explored domain of estimating human pose from signal data.
Finally, we provide future research directions in terms of unsupervised learning, pose representations, and model explainability.
\subsection{Open Questions}
Human pose estimation has been greatly advanced by the deep learning. However, there are still numerous challenges that prevent models from achieving perfect performance.
Such challenges mainly arise from two aspects: question of models and shortcoming of datasets.
\par
\textbf{Model Capacity}\quad
Regarding both \emph{image-based} and \emph{video-based} human pose estimation, modern deep models have difficulties in tackling pose occlusions, person entanglement, and motion blur in complex scenarios.
In such cases, the absence of keypoint visual feature leads to difficulties in localizing joints according to visual information.
For image-based human pose estimation, models require the prior knowledge of human structure to cope with the lack of visual cues in static images.
In terms of video-based human pose estimation, 
the models need to fully use temporal cues to recover human poses from the frames with insufficient visual information.
Additional cues from adjacent frames can be employed to reconstruct the pose of the current frame.
\par
\textbf{Training Data Shortage}\quad
Large-scale annotated image datasets are currently available, yet video datasets still suffer from some shortcomings such as singular scenes and insufficient quantity.
On the other hand, the high-quality position labels of occluded joints are missing in the video dataset. Most of existing video datasets only label the joint visibility to indicate that whether a joint is occluded.
In this configuration, the models are hard to learn to detect the occluded or entangled joints, which greatly increases the difficulty in handling pose occlusions.

In addition, lacking of domain-specific datasets is also a shortcoming.
For particular scenes such as dancing and swimming, datasets of the corresponding domains are necessary for the practical application. 
Therefore, building specialized datasets for various domains is essential to facilitate the application of 2D HPE.
\subsection{Signal-Based Human Pose Estimation}
The corruption of visual features leads to challenges in handling \emph{hard} joints, and non-visual data such as \emph{WIFI signals} provides another way to overcome this problem. 
Previous works \cite{zhao2018through, wang2019can, li2020capturing, wang2019person, guo2019signal} propose to recover human poses from the radio signals or radar.
\cite{zhao2018through} leverages WIFI signals to traverse walls and reflect off the human body for accurately estimating human pose when the person is occluded by the wall.
 Specifically, a deep neural network is proposed to parse keypoint locations from WiFi signals.
\cite{wang2019person} presents a WiFi antennas-based method which takes the WiFi signals as input, and performs pose estimation in an end-to-end fashion.
\cite{li2020capturing} proposes a human pose estimation system using $77GHz$ millimeter wave radar, which first employs two radar data to generate heatmaps, and then employs a CNN to transform two-dimensional heatmaps into human poses.
\vspace{-1em} 
\subsection{Future Directions}
\textcolor{black}{We expect that future researches would dive deeper into three aspects: unsupervised learning, pose representation, and model interpretability.}
\par
\textbf{Unsupervised Learning}\quad
The fully-supervised methods currently dominate the field of human pose estimation since their superior performance.
Their success stems from the rich pose annotations in large-scale datasets.
However, unlabeled images and videos are an almost endless source, and providing full annotations for these data is impossible.
Therefore, unsupervised learning that can automatically learn knowledge of human body from an infinite amount of data has been an important direction.
 \par
 \textbf{Pose Representation}\quad
The heatmap-based pose representation has demonstrated superior performance. However, quantization errors in encoding heatmap from coordinates and decoding coordinates from heatmaps are inevitable.
 Simultaneously, the encoding and decoding processes of the heatmap are influenced by its resolution. 
 The high resolution brings good accuracy, but also increases the computational load.
Therefore, a novel unbiased pose representation for addressing such issues is necessary.
 \par
 \textbf{Model Explainability}\quad
 A drawback of deep learning methods is uninterpretability.
 So far, there is no comprehensive and formal theory for interpretability. 
 As a result, there is limited systematic guidance in designing the deep learning models.
 With respect to human pose estimation, we also fail to clearly understand how the visual features of the input image impact the final keypoint localization, which is detrimental to future investigations. 
Given the potential shortcoming, it is highly desirable to advance works on the interpretability of human pose estimation models.
\section{Conclusion}\label{conclusion}
In this paper, we present a comprehensive and systematic review of human pose estimation methods.
We present a coarse-level taxonomy with three categories: \emph{network architecture design}, \emph{network training refinement}, and \emph{post processing}.
The network architecture design methods focus on the model architecture, the network training refinement methods revolve around the training of networks, and the post processing methods consider the model-agnostic optimization strategies. 
On a finer level, we split the \emph{network architecture design} methods (Section \ref{model}) into top-down framework and bottom-up framework.
We divide the \emph{network training refinement} approaches (Section \ref{train}) into data augmentation techniques, multi-task learning strategies, loss function constraints, and domain adaption methods.
The \emph{post processing} methods (Section \ref{post})  consists of quantization error and pose resampling.
Ultimately, we summarize popular benchmark datasets and evaluation metrics, conduct model performance comparisons, and discuss the potential future  research directions. \textcolor{black}{Hope this would be beneficial for researchers in the community and would inspire future research.}
\section{Acknowledgements}
 This paper is supported by the National Key R\&D Program of China (Grant no.2018YFB1404102), the Key R\&D Program of Zhejiang Province (No. 2021C01104), and the National Natural Science Foundation of China (No. 61902348).

%
%

\bibliographystyle{spbasic}      

\bibliography{References}   

\end{document}